\documentclass[10pt,twocolumn,letterpaper]{article}
\usepackage{cvpr}
\usepackage{times}
\usepackage{graphicx}
\usepackage{epsfig}
\usepackage{epstopdf}
 \pdfoutput=1 
\usepackage{amsmath}
\usepackage{amssymb}
\usepackage{array}
\usepackage{url}
\usepackage{bm}
\usepackage{multirow}
\usepackage[titletoc,title]{appendix}

\usepackage[pagebackref=true,breaklinks=true,letterpaper=true,colorlinks,bookmarks=false]{hyperref}

\cvprfinalcopy 

\def\cvprPaperID{626} 

\begin{document}

\title{A Discriminative CNN Video Representation for Event Detection}
\author{Zhongwen Xu$^\dag$\hspace{4mm} Yi Yang$^\dag$\hspace{4mm} Alexander G. Hauptmann$^\S$\\
$^\dag$ITEE, The University of Queensland, Australia\\
$^\S$SCS, Carnegie Mellon University, USA\\
{\tt\small \{z.xu3, yi.yang\}@uq.edu.au} \hspace{1mm} \hspace{1mm} {\tt\small alex@cs.cmu.edu}
}
\maketitle

\begin{abstract}

	In this paper, we propose a discriminative video representation for event detection over a large scale video dataset when only limited hardware resources are available. The focus of this paper is to effectively leverage deep Convolutional Neural Networks (CNNs) to advance event detection, where only frame level static descriptors can be extracted by the existing CNN toolkit. This paper makes two contributions to the inference of CNN video representation. First, while average pooling and max pooling have long been the standard approaches to aggregating frame level static features, we show that performance can be significantly improved by taking advantage of an appropriate encoding method. Second, we propose using a set of latent concept descriptors as the frame descriptor, which enriches visual information while keeping it computationally affordable. The integration of the two contributions results in a new state-of-the-art performance in event detection over the largest video datasets. Compared to improved Dense Trajectories, which has been recognized as the best video representation for event detection, our new representation improves the Mean Average Precision (mAP) from 27.6\% to 36.8\% for the TRECVID MEDTest~14 dataset and from 34.0\% to 44.6\% for the TRECVID MEDTest~13 dataset.
\end{abstract}

\section{Introduction and Related Work}
Complex event detection~\cite{med13, med14}, which targets the detection of such events as ``renovating a home'' in a large video collection crawled from Youtube, has recently attracted a lot of research attention in computer vision. Compared to concept analysis in videos, \eg, action recognition, event detection is more difficult primarily because an event is more complex and thus has greater intra-class variations. For example, a ``marriage proposal" event may take place indoors or outdoors, and may consist of multiple concepts such as  ring (object), kneeling down (action) and kissing (action).


Recent research efforts have shown that combining multiple features, including static appearance features~\cite{HOG, SIFT, CSIFT}, motion features~\cite{STIP, MoSIFT, dense_trajectories, imtraj, oneata_fv} and acoustic features~\cite{florian}, yields good performance in event detection,
as evidenced by the reports of the top ranked teams in the TRECVID Multimedia Event Detection (MED) competition~\cite{AXES2013, Informedia, Sesame, BBN} and research papers~\cite{BBN_CVPR, Sarnoff, Xu_fusion} that have tackled this problem.
By utilizing additional data to assist complex event detection, researchers propose the use of ``video attributes'' derived from other sources to facilitate event detection~\cite{Ma_multisource}, or to utilize related exemplars when the training exemplars are very few~\cite{Yang_related}. As we focus on improving video representation in this paper, this new method can be readily fed into those frameworks to further improve their performance.


Dense Trajectories and its enhanced version {\em improved Dense Trajectories} (IDT)~\cite{imtraj} have dominated complex event detection in recent years due to their superior performance over other features such as the motion feature STIP~\cite{STIP} and the static appearance feature Dense SIFT~\cite{AXES2013}. Despite good performance, heavy computation costs greatly restrict the usage of the improved Dense Trajectories on a large scale. In the TRECVID MED competition 2014~\cite{med14}, the National Institute of Standards and Technology (NIST) introduced a very large video collection, containing 200,000 videos of 8,000 hours in duration. Paralleling 1,000 cores, it takes about one week to extract the improved Dense Trajectories for the 200,000 videos in the TRECVID MEDEval~14 collection. Even after the spatial re-sizing and temporal down-sampling processing, it still takes 500 cores one week to extract the features~\cite{AXES2013}. As a result of the unaffordable computation cost, it would be extremely difficult for a relatively smaller research group with limited computational resources to process large scale MED datasets. It becomes important to propose an efficient representation for complex event detection with only affordable computational resources, \eg, {\em a single machine}, while at the same time attempting to achieve better performance.

One instinctive idea would be to utilize the deep learning approach, especially  Convolutional Neural Networks (CNNs), given their overwhelming accuracy in image analysis  and fast processing speed, which is achieved by leveraging
the  massive  parallel  processing  power  of GPUs \cite{alexnet}. 
However, it has been reported that the event detection performance of CNN based video representation is worse than the improved Dense Trajectories last year~\cite{Informedia, AXES2013}, as shown in Table~\ref{tb:linear}. A few technical problems remain unsolved.

Firstly, CNN requires a large amount of labeled video data to train good models from scratch.  The large scale TRECVID MED datasets (\ie, MEDTest~13~\cite{med13} and MEDTest~14~\cite{med14}) only have 100 positive examples per event, with many null videos which are irrelevant. The number of labeled videos is smaller than that of the video collection for  sports videos~\cite{deep_video}. In addition, as indicated in \cite{Yang_related}, event videos are quite different from action videos, so it makes little sense to use the action dataset to train models for event detection.




Secondly, when dealing with a domain specific task with a small number of training data, fine-tuning~\cite{rcnn} is an effective technique for adapting the ImageNet pre-trained models for new tasks. However, 
the video level event labels are rather coarse at the frame level, \ie, not all frames necessarily contain the semantic information of the event. If we use the coarse video level label for each frame, performance is barely improved; this was verified by our preliminary experiment.



\begin{table}
\centering
\begin{tabular}{|c|c|c|}
\hline
                                                                                 & MEDTest 13& MEDTest 14\\ \hline
                                                                           IDT~\cite{imtraj, AXES2013}       & \textbf{34.0}      & \textbf{27.6}      \\ \hline \hline
CNN in Lan~\etal~\cite{Informedia} & 29.0 &  N.A. \\ \hline
$\text{CNN}_\text{avg}$ &  32.7 & 24.8 \\\hline
\end{tabular}
\caption{Performance comparison (mean Average Precision in percentage). Lan~\etal~\cite{Informedia} is the only attempt to apply CNN features in TRECVID MED 2013. $\text{CNN}_\text{avg}$ are our results from the average pooling representation of frame level CNN descriptors.}
\label{tb:linear}
\end{table}

Lastly, given the frame level CNN descriptors, we need to generate a discriminative video level representation. Average pooling is the standard approach~\cite{AXES2012, AXES2013} for static local features, as well as for the CNN descriptors~\cite{Informedia}.
Table~\ref{tb:linear} shows the performance comparisons of the improved Dense Trajectories and CNN average pooling representation. We provide the performance of Lan~\etal~\cite{Informedia} for reference as well.
 We can see that the performance of CNN average pooling representation cannot get better than the hand-crafted feature improved Dense Trajectories, which is fairly different from the observations in other vision tasks~\cite{rcnn, deep_eval, Gong}. 

The contributions of this paper are threefold. First, this is the first work to leverage the encoding techniques to generate video representation based on CNN descriptors. Second, we propose to use a set of latent concept descriptors as frame descriptors, which further diversifies the output with aggregation on multiple spatial locations at deeper stage of the network. The approach forwards video frames for only one round along the deep CNNs for descriptor extraction. With these two contributions, the proposed video CNN representation achieves more than 30\% relative improvement over the state-of-the-art video representation on the large scale MED dataset, and this can be conducted on a single machine in two days with a GPU card installed. In addition, we propose to use Product Quantization~\cite{PQ} based on CNN video representation to speed up the execution (event search) time. According to our extensive experiments, we show that  the approach significantly reduces the I/O cost, thereby making event prediction much faster while retaining almost the same level of precision.

\section{Preliminaries}
Unless otherwise specified, this work is based on the network architecture released by~\cite{VGG}, \ie, the configuration with 16 weight layers in the VGG ILSVRC 2014 classification task winning solutions. The first 13 weight layers are convolutional layers, five of which are followed by a max-pooling layer. The last three weight layers are fully-connected layers. In the rest of this paper, we follow the notations in~\cite{deep_eval, rcnn}: $\text{pool}_5$ refers to the activation of the last pooling layer, $\text{fc}_6$ and $\text{fc}_7$ refer to the activation of the first and second fully-connected layers, respectively. Though the structure in~\cite{VGG} is much deeper than the classic CNN structure in~\cite{alexnet, deep_eval, rcnn}, the subscripts of $\text{pool}_5$, $\text{fc}_6$ and $\text{fc}_7$ notations still correspond if we regard the convolution layers between the max-pooling layers as a ``compositional convolutional layer''~\cite{VGG}. We utilize the activations before  Rectified
Linear Units (\ie, $\text{fc}_6$ and $\text{fc}_7$) and after them (\ie, $\text{fc}_6$\_relu and $\text{fc}_7$\_relu), since we observe significant differences in performance between these two variants.

\section{Video CNN Representation}


We begin by 
extracting the frame level CNN descriptors using the Caffe toolkit~\cite{caffe} with the model shared by \cite{VGG}. We then need to generate video level vector representations on top of the frame level CNN descriptors. 

\subsection{Average Pooling on CNN Descriptors}
As described in state-of-the-art complex event detection systems~\cite{AXES2013, AXES2012}, the standard way to achieve image-based video representation in which local descriptor extraction relies on individual frames alone, is as follows: (1) Obtain the descriptors for individual frames; (2) Apply normalization on frame descriptors; (3) {\em Average pooling} on frame descriptors to obtain the video representation, \ie, $\bm{x}_\text{video} = \frac{1}{N} \sum_{i=1}^N \bm{x}_i$, $\bm{x}_i$ is the frame-level descriptor and $N$ is the total number of frames extracted from the video; (4) Re-normalization on video representation. 


Max pooling on frames to generate video representation is an alternative method but it is not typical in event detection. We observe similar performance with average pooling, so we omit this method.

\subsection{Video Pooling on CNN descriptors}

Video pooling computes video representation over the whole video by pooling all the descriptors from all the frames in a video.
The Fisher vector~\cite{FV_ECCV,FV_journal} and Vector of Locally Aggregated Descriptors (VLAD)~\cite{VLAD_CVPR,VLAD_PAMI} have been shown to have great advantages over Bag-of-Words (BoWs)~\cite{video_google} in local descriptor encoding methods. The Fisher vector and VLAD have been proposed for image classification and image retrieval to encode image local descriptors such as dense SIFT and Histogram of Oriented Gradients (HOG). Attempts have also been made to apply Fisher vector and VLAD on local motion descriptors such as Histogram of Optical Flow (HOF) and Motion Boundary Histogram (MBH) to capture the motion information in videos. To our knowledge, {\em this is the first work on the video encoding of CNN descriptors and we broaden the encoding methods from local descriptors to CNN descriptors in video analysis}.
\subsubsection{Fisher Vector Encoding}
In Fisher vector encoding~\cite{FV_ECCV,FV_journal}, a Gaussian Mixture Model (GMM) with $K$ components can be denoted as $\Theta = \{ (\mu_k, \Sigma_k, \pi_k), k=1,2,\ldots, K \}$, where $\mu_k, \Sigma_k, \pi_k$ are the mean, variance and prior parameters of $k$-th component learned from the training CNN descriptors in the frame level, respectively. Given $X = (\bm{x}_1, \ldots, \bm{x}_N)$ of CNN descriptors extracted from  a video, we have mean and covariance deviation vectors for the $k$-th component as:
\begin{align}
\bm{u}_k = \frac{1}{N\sqrt{\pi_k}}  \sum_{i=1}^N q_{ki} \left( \frac{\bm{x}_i - \bm{\mu}_k}{\bm{\sigma}_k} \right) \nonumber  \\ 
\bm{v}_k = \frac{1}{N\sqrt{2\pi_k}} \sum_{i=1}^N q_{ki}  \left[ \left( \frac{\bm{x}_i - \bm{\mu}_k}{\bm{\sigma}_k} \right)^2  - \bm{1} \right],
\end{align}
where $q_{ik}$ is the posterior probability. By concatenation of the $\bm{u}_k$ and $\bm{v}_k$ of all the $K$ components, we form the Fisher vector for the video with size $2D'K$, where $D'$ is the dimension of CNN descriptor $\bm{x}_i$ after PCA pre-processing. PCA pre-processing is necessary for a better fit on the diagonal covariance matrix assumption~\cite{FV_journal}. Power normalization, often Signed Square Root (SSR) with $z = \text{sign}(z) \sqrt{|z|}$,  and $\ell_2$ normalization are then applied to the Fisher vectors~\cite{FV_ECCV, FV_journal}.

\subsubsection{VLAD Encoding}
VLAD encoding~\cite{VLAD_CVPR,VLAD_PAMI} can be regarded as a simplified version of Fisher vector encoding. With $K$ coarse centers $\{\bm{c}_1, \bm{c}_2, \ldots, \bm{c}_K\}$ generated by K-means, we can obtain the difference vector regarding center $c_k$ by:

\begin{equation}
\bm{u}_k = \sum_{i: \text{NN}(\bm{x}_i) = \bm{c}_k} (\bm{x_i} - \bm{c}_k),
\end{equation} 
where $\text{NN}(\bm{x_i})$ indicates $\bm{x}_i$'s nearest neighbors among $K$ coarse centers. 

The VLAD encoding vector with size $D'K$ is obtained by concatenating $\bm{u}_k$ over all the $K$ centers. Another variant of VLAD called VLAD-$k$, which extends the nearest centers with the $k$-nearest centers, has shown good performance in action recognition~\cite{MPEG_flow, Peng_IJCV}. Without specification, we utilize VLAD-$k$ with $k=5$ by default. Except for the power and $\ell_2$ normalization, we apply intra-normalization~\cite{all_VLAD} to VLAD.

\begin{figure}
\centering
\includegraphics[width=0.48\textwidth]{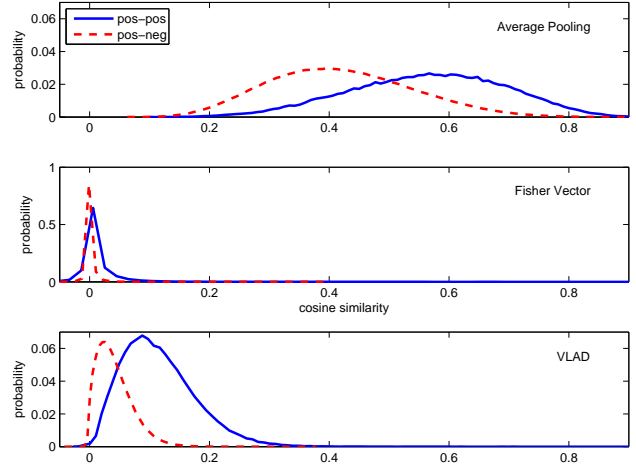}
\caption{Probability distribution of the cosine similarity between positive-positive (blue and plain) and positive-negative (red and dashed) videos using $\text{fc}_7$ features, for average pooling (top), encoding with VLAD using 256 centers (middle), and encoding with the Fisher vector using 256-component GMM (bottom). As the range of probability of Fisher vectors is very different from average pooling and VLAD, we only use consistent axes for average pooling and VLAD. This figure is best viewed in color.}
\label{fig:cosine}
\end{figure}

\begin{figure*}
\centering
\includegraphics[scale=0.4]{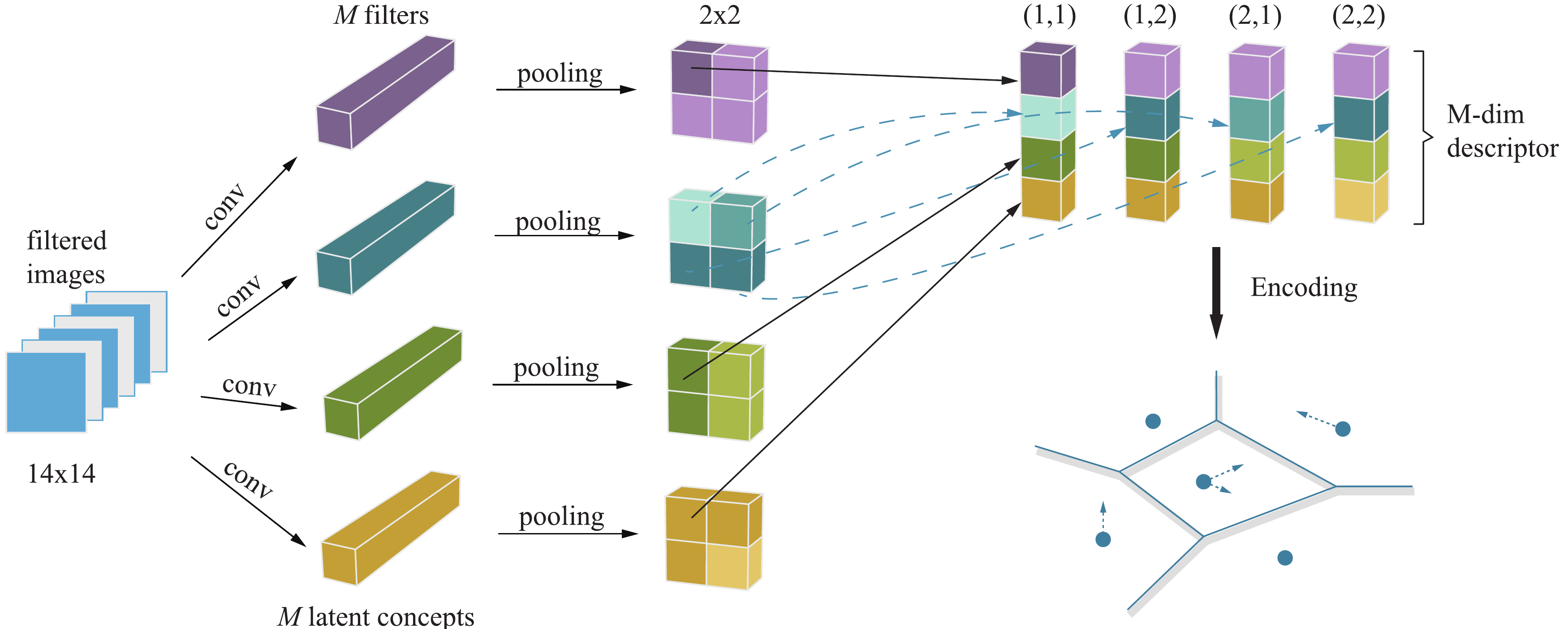}
\caption{Illustration of the latent concept descriptors encoding procedure. We adopt $M$ filters in the last convolutional layer as $M$ latent concept classifiers. Before the last convolutional layer, $M$ filters (\eg, a cuboid of size $3\times3\times512$) produce the prediction outputs at every convolution location, followed by the max-pooling operations. Then, we get the responses of windows of different sizes and strides (in this example the output size is $2\times2$) for each latent concept. Color strength  corresponds to the strength of response of each filter. Finally, we accumulate the responses for the $M$ filters at the same location into the latent concept descriptors. Each dimension corresponds to one latent concept. After obtaining all latent concept descriptors of all frames, we then apply encoding methods to get the final video representation. This figure is best viewed in color.}
\label{fig:latent}
\end{figure*}
\subsubsection{Quantitative Analysis}
\label{sec:quantitative}
Given the above three approaches, we need to find out which one is the most appropriate for the CNN descriptors. 
To this end, we conduct an analytic experiment on the MEDTest~14 training set~\cite{med14} to study the discriminative ability of three types of video representations, \ie, average pooling, video pooling with Fisher vector, and video pooling with VLAD on the CNN descriptors. 
Specifically, we calculate the cosine similarity within the positive exemplars among all the events (denoted as pos-pos), and the cosine similarity between positive exemplars and negative exemplars (denoted as pos-neg). The results are shown in Figure~\ref{fig:cosine}. 
With a good representation,  the data points of positive and negative exemplars should be far away from each other, \ie, the cosine similarity of ``pos-neg" should be close to zero. In addition, there should be a clear difference between the distributions of ``pos-pos" and  ``pos-neg".

\textbf{Average pooling}: In Figure~\ref{fig:cosine}, we observe that the ``pos-neg" cosine similarity distribution is far from zero, which is highly indicative that a large portion of the positive and negative exemplar pairs are similar to each other. In addition, the intersection of areas under the two lines span over a large range of $[0.2, 0.8]$. Both observations imply that average pooling may not be the best choice.

\textbf{Fisher vector}: Although the ``pos-neg" similarity distribution is fairly close to zero, a large proportion of the ``pos-pos" pairs also fall into the same range. No obvious difference between the distributions of ``pos-pos" and ``pos-neg" can be observed.

\textbf{VLAD}: The distribution of the ``pos-neg" pairs is much closer to zero than average pooling while a relatively small proportion of the ``pos-pos" similarity is close to the peak of the ``pos-neg" similarity. 

From the above analytic study, we can see that VLAD is the most fit for the CNN descriptors because the VLAD representation has the best discriminative ability, which is also consistent with the experimental results in Section~\ref{sec:exps}.

\subsection{CNN Latent Concept Descriptors}


Compared to the fully-connected layers, $\text{pool}_5$ contains spatial information. However, if we follow the standard way and flatten $\text{pool}_5$  into a vector, the feature dimension will be very high, which will induce heavy computational cost. 
Specifically, the features dimension of $\text{pool}_5$ is $a \times a \times M$, where $a$ is the size of filtered images of the last pooling layer and $M$ is the number of convolutional filters in the last convolutional layer (in our case, $a = 7$ and $M = 512$). In the VGG network~\cite{VGG}, $\text{pool}_5$ features are vectors of 25,088-D while the $\text{fc}_6$ and $\text{fc}_7$ features have only 4096-D.
As a result, researchers tend to  ignore the general features extracted from $\text{pool}_5$~\cite{deep_eval, Gong}. The problem is even more severe in the video pooling scheme because the frame descriptors with high dimensions would lead to instability problems~\cite{hyper-pooling}.


Note that the convolutional filters can be regarded as generalized linear classifiers on the underlying data patches, and  each convolutional filter corresponds to a latent concept~\cite{NIN}. We propose to formulate the general features from $\text{pool}_5$ as the vectors of {\em latent concept descriptors}, in which each dimension of the latent concept descriptors represents the response of the specific latent concept. Each filter in the last convolutional layer is independent from other filters. The response of the filter is the prediction of the linear classifier on the convolutional location for the corresponding latent concept. In that way, $\text{pool}_5$ layer of size $a\times a \times M$ can be converted into $a^2$ latent concept descriptors with $M$ dimensions. Each latent concept descriptor represents the responses from the $M$ filters for a specific pooling location. Once we obtain the latent concept descriptors for all the frames in a video, we then apply an encoding method to generate the video representation. In this case, each frame contains $a^2$ descriptors instead of  one descriptor for the frame, as illustrated in Figure~\ref{fig:latent}.

In~\cite{SPP}, He~\etal claim that the aggregation at a deeper layer is more compatible with the hierarchical information processing in our brains than cropping or wrapping the original inputs, and they propose to use a Spatial Pyramid Pooling (SPP) layer for object classification and detection, which not only achieves better performance but also relaxes the  constraint that the input must be fixed-size. Different from~\cite{SPP}, we do not train the network with the SPP layer from scratch, because it takes much longer time, especially for a very deep neural network. Instead, at the last pooling layer, we adopt multiple windows with different sizes and strides without retraining the CNNs. In that way, visual information is enriched while only marginal computation cost is added, as we forward frames through the networks only once to extract the latent concept descriptors.

After extracting the CNN latent concept descriptors for all spatial locations of each frame in a video, we then apply video pooling to all the latent concept descriptors of that video. As in~\cite{SPP}, we apply four different CNN max-pooling operations and obtain $(6\times6)$, $(3\times3)$, $(2\times2)$ and $(1\times1)$ outputs for each independent convolutional filter, a total of 50 spatial locations for a single frame. The dimension of latent concept descriptors (512-D) is shorter than the descriptors from the fully-connected layers (4,096-D), while the visual information is enriched via multiple spatial locations on the filtered images.


\subsection{Representation Compression}

For the engineering aspect of a fast event search~\cite{med14} on a large video collection, we can utilize techniques such as Product Quantization (PQ)~\cite{PQ} to compress the Fisher vector or VLAD representation. With PQ compression, the storage space in disk and memory can be reduced by more than an order of magnitude, while the performance remains almost the same. The basic idea of PQ is to decompose the representation into sub-vectors with equal length $B$, and then within each sub-vector, K-means is applied to generate $2^m$ centers as representative points. All the sub-vectors are approximated by the nearest center and encoded into the index of the nearest center. In this way, $B$ float numbers in the original representation become $m$ bit code; thus, the compression ratio is $\frac{B \times 32}{m}$. For example, if we take $m = 8$ and $B = 4$, we can achieve 16 times reduction in storage space. 

Targeting at prediction on compressed data instead of on the original features, we can decompose the learned linear classifier $w$ with an equal length $B$. With look-up tables to store the dot-product between sub-vectors of $2^m$ centers and the corresponding sub-vector of $w$, the prediction speed on large-amount of videos can be accelerated by $\frac{D}{B}$ times look-up operations and $\frac{D}{B} - 1$ times addition operations for each video assuming $D$ is the feature dimension~\cite{FV_journal} .

\section{Experiment Settings}
\subsection{Datasets}
In our experiments, we utilize the largest event detection datasets with labels\footnote{Labels for MEDEval 13 and MEDEval 14 are not publicly available.}, namely TRECVID MEDTest 13~\cite{med13} and TRECVID MEDTest 14~\cite{med14}. They have been introduced by NIST for all participants in the TRECVID competition and research community to conduct experiments on. For both datasets, there are 20 complex events respectively, but with 10 events overlapping. MEDTest~13 contains events E006-E015 and E021-E030, while MEDTest~14 has events E021-E040. Event names include ``Birthday party", ``Bike trick", \etc. Refer to~\cite{med13,med14} for the complete list of event names. In the training section, there are approximately 100 positive exemplars per event, and all events share negative exemplars with about 5,000 videos. The testing section has approximately 23,000 search videos. The total duration of videos in each collection is about 1,240 hours. 
\subsection{Features for Comparisons}

As reported in \cite{AXES2013} and compared with the features from other top performers~\cite{BBN, Sesame, Informedia} in the TRECVID MED 2013 competition, we can see that the improved Dense Trajectories has superb advantages over the original Dense Trajectories (used by all other teams except~\cite{AXES2013}), and is even better than approaches that combine many low-level visual features~\cite{BBN, Sesame, Informedia}.
Improved Dense Trajectories extracts local descriptors such as trajectory, HOG, HOF, and MBH, and Fisher vector is then applied to encode the local descriptors into video representation. Following~\cite{imtraj, AXES2013}, we first reduce the dimension of each descriptor by a factor of 2 and then utilize 256 components to generate the Fisher vectors. We evaluate four types of descriptor in improved Dense Trajectories, and report the results of the best combination of descriptors and the two individual descriptors that have the best performance (HOG and MBH). 

In addition, we report the results of some popular features used in the TRECVID competition for reference, such as STIP~\cite{STIP}, MoSIFT~\cite{MoSIFT} and CSIFT~\cite{CSIFT}, though their performance is far weaker than improved Dense Trajectories.

\subsection{Evaluation Details}
In all the experiments, we apply linear Support Vector Machine (SVM) with LIBSVM toolkit~\cite{libsvm}. We conduct extensive experiments on two standard training conditions: in 100Ex, 100 positive exemplars are given in each event and in 10Ex, 10 positive exemplars are given. In the 100Ex condition, we utilize 5-fold cross-validation to choose the parameter of regularization coefficient $C$ in linear SVM. In the 10Ex condition, we follow~\cite{Informedia} and set $C$ in linear SVM to 1.


We sample every five frames in the videos and follow the pre-processing of~\cite{alexnet, deep_eval} on CNN descriptor extraction. We extract the features from the center crop only. CNN descriptors are extracted using Caffe~\cite{caffe} with the best publicly available model~\cite{VGG}, and we utilize vlfeat~\cite{vlfeat} to generate Fisher vector and VLAD representation.

Mean Average Precision (mAP) for binary classification is applied to evaluate the performance of event detection according to the NIST standard~\cite{med13, med14}. 
\section{Experiment Results}
\subsection{Results for Video Pooling of CNN descriptors}
\label{sec:exps}
In this section, we show the experiments on video pooling of $\text{fc}_\text{6}$, $\text{fc}_\text{6}$\_relu, $\text{fc}_\text{7}$ and $\text{fc}_\text{7}$\_relu.
Before aggregation, we first apply PCA with whitening on the $\ell_2$ normalized CNN descriptors. Unlike local descriptors such as HOG, MBH, which have dimensions less than 200-D, the CNN descriptors have much higher dimensions (4,096-D). We conduct experiments with different reduced dimensions, \ie, 128, 256, 512 and 1,024, and utilize the reduced dimensions that best balance performance and storage cost in corresponding features, \ie, 512-D for $\text{fc}_6$ and $\text{fc}_6$\_relu and 256-D for $\text{fc}_7$ and $\text{fc}_7$\_relu. We utilize 256 components for Fisher vectors  and 256 centers for VLAD as common choices in~\cite{FV_journal, VLAD_CVPR}. We will study the impact of parameters in Section~\ref{sec:params}. PCA projections, components in GMM for Fisher vectors, and centers in K-means for VLAD are learned from approximately 256,000 sampled frames in the training set. 

Since we observe similar patterns in MEDTest~13 and MEDTest~14 under both 100Ex and 10Ex, we take MEDTest~14 100Ex as an example to compare with different representations, namely average pooling, video pooling with Fisher vectors and video pooling with VLAD. From Table~\ref{tb:FV_VLAD}, we can see that both video pooling with Fisher vectors and VLAD demonstrate great advantages over the average pooling representation. On the video pooling of CNN descriptors, Fisher vector encoding does not exhibit better performance than VLAD. Similar observations have been expressed in~\cite{hyper-pooling}. We suspect that the distribution of CNN descriptors is quite different from the local descriptors, \eg, HOG, HOF. We will study the theoretical reasons for the poorer performance of Fisher vector than VLAD on CNN video pooling in future research.


\begin{table}[h]
\centering
\begin{tabular}{|c|c|c|c|c|}
\hline
              & $\text{fc}_6$   & $\text{fc}_6$\_relu & $\text{fc}_7$   & $\text{fc}_7$\_relu \\ \hline
Average pooling & 19.8 & 24.8 & 18.8 & 23.8 \\ \hline
Fisher vector & 28.3 &         28.4  &   27.4    &    29.1       \\ \hline
VLAD          & \textbf{33.1} &   32.6   & \textbf{33.2} &   31.5   \\ \hline
\end{tabular}
\caption{Performance comparison (mAP in percentage) on \textbf{MEDTest~14 100Ex}}
\label{tb:FV_VLAD}
\end{table}

\begin{figure}
\centering
\includegraphics[width=0.48\textwidth]{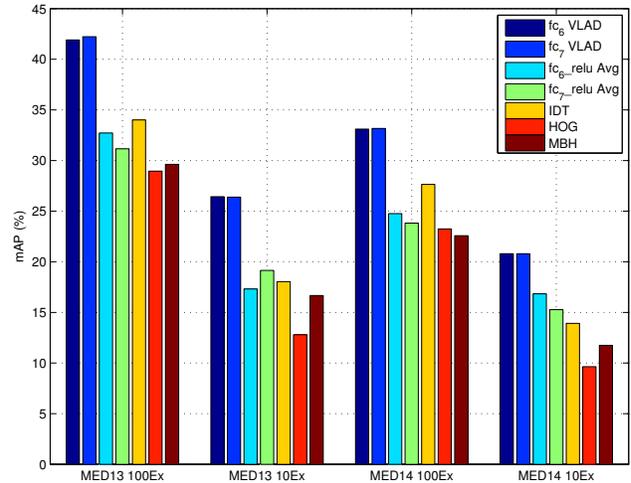}
\caption{Performance comparisons on MEDTest~13 and MEDTest~14, both 100Ex and 10Ex. This figure is best viewed in color.}
\label{fig:VLAD}
\end{figure}

We compare the performance of VLAD encoded CNN descriptors with state-of-the-art feature improved Dense Trajectories (IDT) and average pooling on CNN descriptors in Figure~\ref{fig:VLAD}. We also illustrate the performance of the two strongest descriptors inside IDT (HOG and MBH). We can see very clearly that VLAD encoded CNN features significantly outperform IDT and average pooling on CNN descriptors over all settings. For more references, we provide the performance of a number of widely used features~\cite{Sesame, BBN, Informedia} on MEDTest 14 for comparison. MoSIFT~\cite{MoSIFT} with Fisher vector achieves mAP 18.1\% on 100Ex and 5.3\% on 10Ex; STIP~\cite{STIP} with Fisher vector achieves mAP 15.0\% on 100Ex and 7.1\% on 10Ex; CSIFT~\cite{CSIFT} with Fisher vector achieves mAP 14.7\% on 100Ex and 5.3\% on 10Ex. Note that with VLAD encoded CNN descriptors, we can achieve better performance with 10Ex than the relatively poorer features such as MoSIFT, STIP, and CSIFT with 100Ex!



\subsection{Results for CNN Latent Concept Descriptors with Spatial Pyramid Pooling}
We evaluate the performance of {\em latent concept descriptors} (LCD) of both the original CNN structure and the structure with the Spatial Pyramid Pooling (SPP) layer plugged in to validate the effectiveness of SPP. Before encoding the latent concept descriptors, we first apply PCA with whitening. Dimension reduction is conducted from 512-D to a range of dimensions such as 32-D, 64-D, 128-D, and 256-D, and we find that 256-D is the best choice. We observe a similar pattern with video pooling of fc layers indicating that Fisher vector is inferior to VLAD on video pooling. We omit the results for Fisher vector due to limited space.


\begin{table}[h]
\centering
\begin{tabular}{|c|c|c|}
\hline
                & 100Ex & 10Ex \\ \hline
Average pooling & 31.2  & 18.8 \\ \hline
$\text{LCD}_\text{VLAD}$        & 38.2  & 25.0 \\ \hline
$\text{LCD}_\text{VLAD}$ + SPP            & \textbf{40.3}  & \textbf{25.6} \\ \hline
\end{tabular}
\caption{Performance comparisons for $\text{pool}_5$ on \textbf{MEDTest~13}. $\text{LCD}_\text{VLAD}$ is VLAD encoded LCD from the original CNN structure, while $\text{LCD}_\text{VLAD}$ + SPP indicates VLAD encoded LCD with SPP layer plugged in.}
\label{tb:LCD13}
\end{table}

\begin{table}[h]
\centering
\begin{tabular}{|c|c|c|}
\hline
                & 100Ex & 10Ex \\ \hline
Average pooling & 24.6  & 15.3 \\ \hline
$\text{LCD}_\text{VLAD}$        & 33.9  & 22.8 \\ \hline
$\text{LCD}_\text{VLAD}$ + SPP   & \textbf{35.7}  & \textbf{23.2} \\ \hline
\end{tabular}
\caption{Performance comparisons for $\text{pool}_5$ on \textbf{MEDTest~14}. Notations are the same as Table~\ref{tb:LCD13}.}
\label{tb:LCD14}
\end{table}

We show the performance of our proposed latent concept descriptors (LCD) in Table~\ref{tb:LCD13} and Table~\ref{tb:LCD14}. In both 100Ex and 10Ex over two datasets, we can see clear gaps over the $\text{pool}_5$ features with average pooling, which demonstrates the advantages of our proposed novel utilization of $\text{pool}_5$. With SPP layer, VLAD encoded LCD  ($\text{LCD}_\text{VLAD}$ + SPP) continues to increase the performance further from the original structure ($\text{LCD}_\text{VLAD}$). The aggregation at a deeper stage to generate multiple levels of spatial information via multiple CNN max-pooling demonstrates advantages over the original CNN structure while having only minimal computation costs. The SPP layer enables a single pass of the forwarding in the network compared to the multiple passes of applying spatial pyramid on the original input images.



\subsection{Analysis of the Impact of Parameters}
\label{sec:params}
We take VLAD encoded $\text{fc}_7$ features under MEDTest 14 100Ex as an example to see the impact of parameters in the video pooling process.

\textbf{Dimensions of PCA}:
The original dimension of $\text{fc}_7$ is quite high compared to local descriptors. It is essential to investigate the impact of dimensions in PCA in the pre-processing stage, since it is critical to achieve a better trade-off of performance and storage costs. Table~\ref{tb:pca} shows that in dimensions of more than 256-D, performance remains similar, whereas encoding in 128-D damages the performance significantly.

\begin{table}[h]
\centering
\begin{tabular}{|c|c|c|c|c|}
\hline
Dimension & 128-D  & 256-D  & 512-D  & 1024-D \\ \hline
mAP       & 30.6 & \textbf{33.2} & 33.1 & 33.2 \\ \hline
\end{tabular}
\caption{Impact of dimensions of CNN descriptors after PCA, with fixed $K = 256$ in VLAD.}
\label{tb:pca}
\end{table}
\textbf{Number of Centers in Encoding}: We explore various numbers of centers $K$ in VLAD, and the results are shown in Table~\ref{tb:num_centers}. With the increase of $K$, we can see that the discriminative ability of the generated features improves. However when $K = 512$, the generated vector may be too sparse, which is somewhat detrimental to performance.


\begin{table}[h]
\centering
\begin{tabular}{|c|c|c|c|c|c|}
\hline
$K$ & 32   & 64   & 128  & 256 & 512  \\ \hline
mAP               & 28.7 & 29.7 & 30.4 & \textbf{33.2} & 32.1 \\ \hline
\end{tabular}
\caption{Impact on numbers of centers ($K$) in VLAD, with fixed PCA dimension of 256-D.}
\label{tb:num_centers}
\end{table}
\textbf{VLAD-$k$}: We experiment with the traditional VLAD as well, with nearest center only instead of $k$-nearest centers. mAP drops from 33.2\% to 32.0\%. 

\textbf{Power Normalization}:
We remove the SSR post-processing and test the features on the VLAD encoded $\text{fc}_7$. mAP drops  from 33.2\% to 27.0\%, from which we can see the significant effect of SSR post-processing.

\textbf{Intra-normalization}: We turn off the intra-normalization. mAP drops from 33.2\% to 30.6\%.


\subsection{Results for Product Quantization Compression}
\begin{table}[h]
\centering
\begin{tabular}{|c|c|c|c|}
\hline
&original & $B=4$  & $B=8$\\ \hline
mAP & 33.2    &         33.5 ($\uparrow 0.3$)        &     33.0 ($\downarrow 0.2$)             \\ \hline
space reduction & - & 16$\times$ & 32$\times$ \\ \hline
\end{tabular}
\caption{Performance change analysis for VLAD encoded $\text{fc}_7$ with PQ compression. $B$ is the length of the sub-vectors in PQ and $m=8$.}
\label{tb:PQ}
\end{table}
We conduct experiments on VLAD encoded $\text{fc}_7$ to see the performance changes with Product Quantization (PQ) compression. From the results in Table~\ref{tb:PQ}, we can see that PQ with $B=4$ maintains the performance and even improves slightly. When $B=8$, performance drops slightly. If we compress with $B=4$ , we can store VLAD encoded $\text{fc}_7$ features  in 3.1 GB for the MEDEval 14, which contains 200,000 videos of 8,000 hours' duration. With further compression with a lossless technique such as Blosc\footnote{Blosc can reduce the storage space by a factor of 4}\cite{inria_blosc}, we can store the features of the whole collection in less than 1 GB, which can be read by a normal SSD disk in a few seconds. Without PQ compression, the storage size of the features would be 48.8 GB, which severely compromises the execution time due to the I/O cost. Utilization of compression techniques largely saves the I/O cost in the prediction procedure, while preserving the performance.


In our speed test on the MEDEval~14 collection using the compressed data but not the original features, we can finish the prediction on 200,000 videos in 4.1 seconds per event using 20 threads on an Intel Xeon E5-2690v2 @ 3.00 GHz.
\subsection{Results for Fusing Multiple Layers Extracted from the Same Model}
We investigate average late fusion~\cite{late_fusion} to fuse the prediction results from different layers with PQ compression, \ie, VLAD encoded LCD with SPP, $\text{fc}_6$ and $\text{fc}_7$. From Table~\ref{tb:final} we can see that the simple fusion pushes the performance further beyond the single layers on MEDTest 13 and MEDTest 14, and achieves significant advantages over improved Dense Trajectories (IDT). Our proposed method pushes the state-of-the-art performance much further, achieves {\em more than 30\% relative improvement} on 100Ex, and {\em more than 65\% relative improvement} on 10Ex over both challenging datasets.

\begin{table}[h]
\centering
\begin{tabular}{|l|c|c|c|}
\hline
                 & Ours & IDT   &  Relative Improv \\ \hline
MED13~100Ex & \textbf{44.6}       & 34.0    &  31.2\% \\ \hline
MED13~~10Ex  & \textbf{29.8}         & 18.0 &   65.6\% \\ \hline
MED14~100Ex & \textbf{36.8}         & 27.6    &  33.3\% \\ \hline
MED14~~10Ex  & \textbf{24.5}         & 13.9   &  76.3\% \\ \hline
\end{tabular}
\caption{Performance comparison of all settings; the last column shows the relative improvement of our proposed representation over IDT.}
\label{tb:final}
\end{table}

\begin{figure}
\centering
\includegraphics[width=0.5\textwidth]{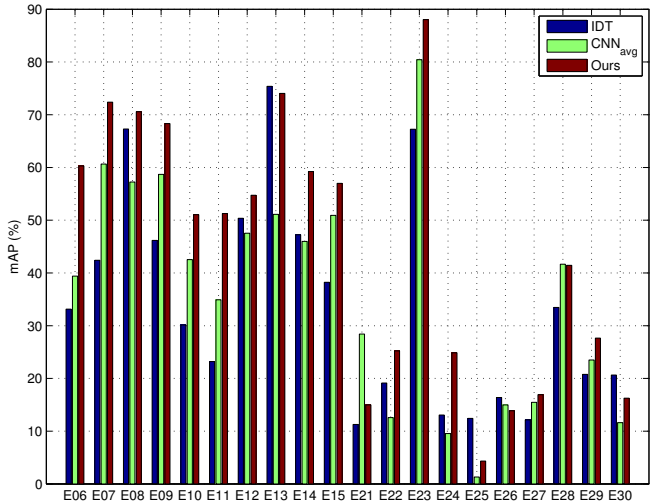}
\caption{MEDTest 13 100Ex per event performance comparison (in mAP percentage). This figure is best viewed in color.}
\label{fig:MED13_per_event}
\end{figure}

\begin{figure}
\centering
\includegraphics[width=0.5\textwidth]{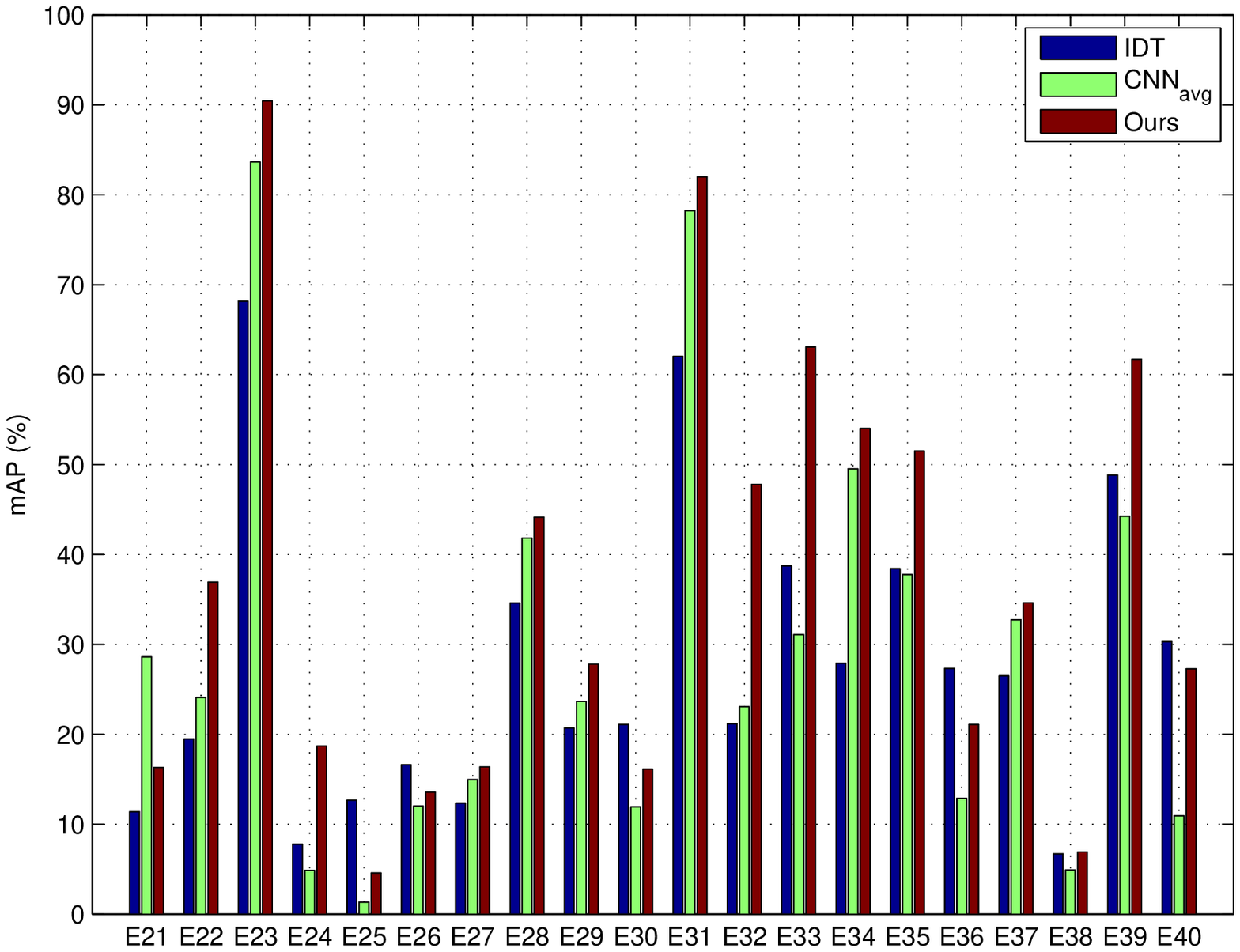}
\caption{MEDTest 14 100Ex per event performance comparison (in mAP percentage). This figure is best viewed in color.}
\label{fig:MED14_per_event}
\end{figure}

Figure~\ref{fig:MED13_per_event} and Figure~\ref{fig:MED14_per_event} show the per-event mAP comparison of the 100Ex setting on MEDTest 13 and MEDTest 14. We provide results for average pooling on CNN descriptors with late fusion of three layers as well, denoted as $\text{CNN}_\text{avg}$. Our proposed representation beats two other strong baselines in {\em 15 out of 20 events} in MEDTest 13 and {\em 14 out of 20 events} in MEDTest 14, respectively.

\subsection{Comparison to the state-of-the-art Systems}
We compare the MEDTest 13\footnote{In~\cite{AXES2013,BBN,Informedia}, teams report performance on MEDEval 13 as well, while MEDEval 13 is a different collection used in the competition, where only NIST can evaluate the performance.} results with the top performers in the TRECVID MED 2013 competition~\cite{AXES2013, BBN, Informedia}. The AXES team does not show their performance on MEDTest~13~\cite{AXES2013}. Natarajan~\etal~\cite{BBN} report mAP 38.5\% on 100Ex, 17.9\% on 10Ex from their {\em whole visual system} of combining all their low-level visual features. Lan~\etal~\cite{Informedia} report 39.3\% mAP on 100Ex of their {\em whole system including non-visual features} while they conducted 10Ex on their internal dataset. Our results achieve 44.6\% mAP on 100Ex and 29.8\% mAP on 10Ex, which significantly outperforms the top performers in the competition who combine more than 10 kinds of features with sophisticated schemes. To show that our representation is complementary to features from other modalities, we perform average late fusion of our proposed representation with IDT and MFCC, and generate a lightweight system with static, motion and acoustic features, which achieves 48.6\% mAP on 100Ex, and 32.2\% mAP on 10Ex. When the reports for TRECVID MED 2014 are available, we will also compare the MEDTest 14 performance with the top performers.


\section{Conclusion}
TRECVID Multimedia Event Detection (MED) has suffered from huge computation costs in feature extraction and classification processes. Using Convolutional Neural Network (CNN) representation seems to be a good solution, but generating video representation from CNN descriptors has different characteristics from image representation. We are the first to leverage encoding techniques to generate video representation from CNN descriptors. And we propose latent concept descriptors to generate CNN descriptors more properly. For fast event search, we utilize Product Quantization to compress the video representation and predict on the compressed data. Extensive experiments on the two largest event detection collections under different training conditions demonstrate the advantages of our proposed representation. We have achieved promising performance which is superior to the state-of-the-art systems which combine 10 more features. The proposed representation is extendible and the performance can be further improved by better CNN models and/or appropriate fine-tuning techniques.


{\small
\bibliographystyle{ieee}
\bibliography{MED_CNN}
}

\appendix
\appendixpage
\section{Non-linear Classifiers on CNN Descriptors}
Noting the observation in~\cite{rcnn} that performance improvement from fine-tuning is much larger for $\text{fc}_6$ and $\text{fc}_7$ than for $\text{pool}_5$, \cite{rcnn} suggests that {\em most of the improvement of fine-tuning is gained from learning domain-specific non-linear classifiers on top of them}. 


We take the spirit of the neural network fine-tuning technique, and show that in our experiments, {\em non-linear classifiers} such as exponential-$\chi^2$ kernel SVM or RBF kernel SVM on top of the CNN descriptors can boost the video classification performance significantly over the standard linear approach~\cite{deep_eval, rcnn}. Exponential-$\chi^2$ kernel and RBF kernel have been investigated in various applications of visual recognition area with the hand-crafted features such as SIFT, HOG. In deep learning era, exponential-$\chi^2$ kernel and RBF kernel can still show performance advantages over linear kernel.

The kernel functions $K(X_i, X_j)$ of exponential-$\chi^2$ kernel and RBF kernel between two data points $X_i$ and $X_j$ are generalized as below,

\begin{equation}
 K(X_i, X_j) = \exp\left( - \frac{1}{A \sigma^2} \text{Dist}(X_i, X_j) \right),
\end{equation}
where $X_i = \{x_{id}\}$ and $X_j = \{x_{jd}\}$ are the deep learning features extracted from intermediate layers in CNNs, \eg $\text{pool}_5$, $\text{fc}_6$, $\text{fc}_7$. $A$ is the average distances (with corresponding distance metric) between all the training features, and  $\sigma$ is the parameter for the kernel function. By utilizing different distance metric we can obtain different kernel. For exponential-$\chi^2$ kernel ,  $\text{Dist}(X_i, X_j)$ is $\chi^2$ distance between $X_i$ and $X_j$, defined as:
\begin{equation}
\text{Dist}_{\chi^2}(X_i, X_j) = \frac{1}{2} \sum_{d = 1}^D \frac{ (x_{id} - x_{jd})^2 } {x_{id} + x_{jd} + \epsilon},
\end{equation}
with $D$ as the dimension of features $X_i, X_j$, and a small value $\epsilon$ to avoid ``divided by zero", while for RBF kernel, $\text{Dist}(X_i, X_j)$ is Euclidean distance between $X_i$ and $X_j$.
\begin{equation}
\text{Dist}_{\text{RBF}}(X_i, X_j) = \frac{1}{2} \sum_{d=1}^{D} (x_{id} - x_{jd})^2,
\end{equation}

Though we show great performance on this simple idea of utilizing non-linear classifiers on CNN descriptors after average pooling in, non-linearity makes it hard to be applied on large-scale event detection like the MEDEval 14 collection with 200,000 videos. We provide the appendix here for improving the performance of average pooling on CNN descriptors at the scale of MEDTest 13 and MEDTest 14.

\section{Experiment Results for Non-linear Classifiers} 
\label{sec:non_linear}
We conduct the experiments with the same settings of the main paper but with kernelized SVM using LIBSVM toolkit~\cite{libsvm}. The additional parameter $\sigma$ for kernel function computation is chosed by 5-fold cross-validation in 100Ex but fixed to 1 in 10Ex.
\begin{table}[h]
\small
\centering
\begin{tabular}{|c|c|c|c|c|c|}
\hline
            & $\text{pool}_5$           & $\text{fc}_6$             & $\text{fc}_6$\_relu       & $\text{fc}_7$            & $\text{fc}_7$\_relu       \\ \hline
linear      & 31.2          & 27.8          & 32.7          & 26.4         & 32.0          \\ \hline
RBF         & 34.4          & 36.6          & 37.9          & 34.9      & 37.2          \\ \hline
$\exp \chi^2$ & \textbf{36.1} & \textbf{38.3} & \textbf{38.9} & \textbf{38.1} & \textbf{39.2} \\ \hline
\end{tabular}
\caption{Performance comparison (mAP, in percentage) with different kernels in \textbf{MEDTest 13 100Ex}, {\em IDT is with \textbf{34.0} mAP}}
\label{tb:MEDTest13_100}
\end{table}

\begin{table}[h]
\small
\centering
\begin{tabular}{|c|c|c|c|c|c|}
\hline
              & $\text{pool}_5$           & $\text{fc}_6$             & $\text{fc}_6$\_relu      & $\text{fc}_7$             & $\text{fc}_7$\_relu      \\ \hline
linear        & 18.8          & 16.0          & 20.7         & 17.3          & 19.2         \\ \hline
RBF           & 20.7          & 20.2          & 24.0         & 20.7          & 22.7         \\ \hline
$\exp \chi^2$ & \textbf{22.2} & \textbf{23.8} & \textbf{24.5} & \textbf{22.8} & \textbf{24.3} \\ \hline
\end{tabular}
\caption{Performance comparison (mAP, in percentage) with different kernels in \textbf{MEDTest 13 10Ex}, {\em IDT is with \textbf{18.0} mAP}}
\label{tb:MEDTest13_10}
\end{table}

\begin{table}[h]
\small
\centering
\begin{tabular}{|c|c|c|c|c|c|}
\hline
         & $\text{pool}_5$  & $\text{fc}_6$    & $\text{fc}_6$\_relu & $\text{fc}_7$    & $\text{fc}_7$\_relu \\ \hline
linear   & 24.6 & 19.9 & 24.8    & 18.8 & 23.8    \\ \hline
RBF      & 28.6 & 29.2 & 31.4    & 29.1 & 29.2    \\ \hline
$\exp\chi^2$ & \textbf{30.8} & \textbf{30.8} & \textbf{32.2}    & \textbf{30.8} & \textbf{32.1}    \\ \hline
\end{tabular}
\caption{Performance comparison (mAP, in percentage) with different kernels in \textbf{MEDTest 14 100Ex}. {\em IDT is with \textbf{27.6} mAP}.}
\label{tb:MEDTest14_100}
\end{table}

\begin{table}[h]
\small
\centering
\begin{tabular}{|c|c|c|c|c|c|}
\hline
              & $\text{pool}_5$           & $\text{fc}_6$             & $\text{fc}_6$\_relu       & $\text{fc}_7$             & $\text{fc}_7$\_relu       \\ \hline
linear        & 15.3          & 12.1          & 16.8          & 13.5          & 15.3          \\ \hline
RBF           & 18.1          & 16.3          & 20.3           & 16.6           & 18.8          \\ \hline
$\exp \chi^2$ & \textbf{19.9} & \textbf{20.0} & \textbf{20.6} & \textbf{20.2} & \textbf{19.8} \\ \hline
\end{tabular}
\caption{Performance comparison (mAP, in percentage) with different kernels in \textbf{MEDTest 14 10Ex}, {\em IDT is with \textbf{13.9} mAP}}
\label{tb:MEDTest14_10}
\end{table}

Here we show the performance of event detectors trained from average pooled CNN descriptors from key frames on MEDTest 13 (Table~\ref{tb:MEDTest13_100} for 100Ex and Table~\ref{tb:MEDTest13_10} for 10Ex) and MEDTest 14 (Table~\ref{tb:MEDTest14_100} for 100Ex and Table~\ref{tb:MEDTest14_10} for 10Ex). We attach the performance of improved Dense Trajectories on each setting for comparison as well. We can see that non-linear classifiers effectively boost the performance. It makes a clear gap ahead hand-crafted features, while it keeps the low-dimensionality advantages as well.


Looking into the details of the performance, we can see (1) Both RBF and exponential-$\chi^2$ non-linear classifiers are significantly better than the linear classifier. Exponential-$\chi^2$ kernel is observably better than RBF kernel in all the layers from all the settings. By applying exponential-$\chi^2$ SVM, the performance of CNN features is \textbf{on average 5\% absolute mAP higher than the state-of-the-art features improved Dense Trajectories}. (2) In linear space, the performance of fully-connected layers with ReLU neurons significantly outperforms the one without ReLU neurons (except the MEDTest 14 10Ex case), which is consistent with the common choice of layers in previous papers~\cite{rcnn}; (3) $\text{pool}_5$ with much higher dimensions do not show advantages over fc layers. Thus, we suggest that {\em when applying event detection on  the average pooled CNN descriptors, we should use features extracted from $\text{fc}_6$\_relu and $\text{fc}_7$\_relu, and then apply exponential-$\chi^2$ kernel classifier to achieve good performance}. 

Though non-linear classifiers can achieve much better performance than the linear classifiers for complex event detection, it cannot satisfy the efficiency requirement in the large-scale application. For non-linear classifiers, when a new test exemplar comes, it has to calculate the kernel matrix between the test exemplar and all the training exemplars, so that we can apply classifiers on the data. In the linear classifier case, we only need to conduct dot-product operation between the learned classifier and the test exemplar, which is thousands times faster in our complex event detection case. 


To tackle this problem, people come up approaches with approximation of the kernel matrix like explicit feature mapping~\cite{efm}.   In our experiments, the explicit feature mapping approximation leads to about 2\% absolute mAP drop with three times dimension as the original features, which is consistently with performance drop of the same approach in other papers~\cite{chebyshev}. The performance drop is mainly due to the fact that the explicit feature mapping is approximating $\chi^2$ kernel, while $\chi^2$ kernel has been shown inferior to exponential-$\chi^2$ kernels in various vision tasks~\cite{chebyshev}.

\end{document}